\documentclass[final,1p,times,authoryear]{elsarticle}

\graphicspath{ {./images/} }

\usepackage{amssymb}

\usepackage{amsmath}

\journal{Neural Networks}

\begin{document}

\begin{frontmatter}



\title{Arithmetic with Language Models: from Memorization to Computation
\tnoteref{labelAccepted}
}
\tnotetext[labelAccepted]{The article has been accepted for publication in Elsevier Neural Networks journal. The final version is available on the Elsevier ScienceDirect platform.}


\author[1]{Davide Maltoni}

\author[1]{Matteo Ferrara}

\affiliation[1]{organization={Department of computer science and engineering , University of Bologna},
            country={Italy}}

\begin{abstract}
A better understanding of the emergent computation and problem-solving capabilities of recent large language models is of paramount importance to further improve them and broaden their applicability. This work investigates how a language model, trained to predict the next token, can perform arithmetic computations generalizing beyond training data. Binary addition and multiplication constitute a good testbed for this purpose, since they require a very small vocabulary and exhibit relevant input/output discontinuities making smooth input interpolation ineffective for novel data. We successfully trained a light language model to learn these tasks and ran a number of experiments to investigate the extrapolation capabilities and internal information processing. Our findings support the hypothesis that the language model works as an Encoding-Regression-Decoding machine where the computation takes place in the value space once the input token representation is mapped to an appropriate internal representation.
\end{abstract}

\begin{keyword}
Language Models \sep AI Explainability \sep Probing \sep Interpretability \sep Arithmetic


\end{keyword}

\end{frontmatter}


\section{Introduction}

Large Language Models (LLMs) based on Transformer architecture \citep{Vaswani2017} have recently demonstrated surprising problem-solving capabilities that require logic reasoning, advanced information processing and common sense \citep{Bubeck2023, Wei2023,Wei2022}. Their huge storage capacity combined with a massive training on terabytes of heterogeneous data could suggest that the memorization of an enormous amount of knowledge is enough to perform well on similar test data. However, validations on carefully selected Out-of-Distribution (OoD) data proved their reasoning capabilities on novel examples requiring non-trivial generalizations. Unfortunately, the depth and width of such models is so high that decoding and understanding the internal information processing is very challenging.

Focusing on arithmetic calculations, some studies \citep{Yuan2023} demonstrate that recent LLMs (such as GPT-4) can perform additions and multiplications with long-digit operands, for which the number of variants is so high to exclude the exhaustive memorization of the training set. Nevertheless, the computational approach put in place by LLMs, as well as the interpolation/extrapolation capabilities remain unexplained.

In this work we design some controlled experiments, consisting of simple computation tasks such as binary addition and multiplication, and solve them with two Language Models (LMs) based on Transformer architecture: (i) the original encoder-decoder architecture by \citet{Vaswani2017} and (ii) a more recent decoder-only architecture denoted as nanoGPT \citep{Karpathy2022}. In spite of their simplicity, these tasks cannot be solved by pure memorization or smooth interpolation and investigating how an LM learn them can improve our understanding of the underlying mechanisms. In particular, using a tiny vocabulary of just 5 tokens and a small training set allows to operate with a light (non-pretrained) LM and use interpretability techniques to investigate internal information processing.

Other studies addressed the ability of LLMs to perform arithmetic computation and train small LMs to learn these tasks from scratch (see related works in Section \ref{sec:RelatedWorks}). However, our aim is different: we are not interested in finding the best LM architecture and setup to maximize accuracy on arithmetic operations, but we look for a simple architecture and setup that allow to effectively solve the task in order to be able to investigate the underlying computational approach. The main novelty and contribution of this work are the formalization of the hypothesis that our LM works as an Encoding-Regression-Decoding machine and the design of a number of experiments to support and validate this hypothesis (see Table \ref{table:InvestigationSteps}).

After presentation of related works in Section \ref{sec:RelatedWorks}, in Section \ref{sec:ExperimentDesign} we introduce the experimental testbed and the architecture of the LM used. Section \ref{sec:Results} presents the results achieved and introduces control experiments and elaborations to shed light on the computation approach used to solve the tasks. In Section \ref{sec:AblationStudy} an ablation study is presented and, finally, in Section \ref{sec:Conclusions} we include a final discussion and draw some conclusions.

\begin{table}[h!]
\caption{The main contributions of this work.}
\label{table:InvestigationSteps}
\begin{center}
\begin{tabular}{ |p{5cm}|p{5cm}|c| } 
 \hline
 \textbf{Step} & \textbf{Aim} & \textbf{Where} \\
 \hline
 Training an LM on addition and multiplication & Demonstrating these arithmetic problems can be solved with a simple LM trained form scratch & Section \ref{subsec:LearningAdditionAndMultiplication} \\
 \hline
 Manipulating training set by excluding specific regions of the input space & Evaluating interpolation/extrapolation capabilities and making hypothesis on internal regression & Section \ref{subsec:InterpolationVsExtrapolation} \\
 \hline
 Correlation analysis of internal values (embeddings) & Support hypothesis that LM works as an ERD machine & Section \ref{subsec:LookingAtInternalRepresentations} \\
 \hline
 Amnesic probing & Prove that the “value” information is crucial to properly compute the output & \ref{sec:AppendixD} \\
 \hline
\end{tabular}
\end{center}
\end{table}

\section{Related works} \label{sec:RelatedWorks}

\subsection{LM and LLM capabilities on arithmetic tasks}

In \citet{Yuan2023} recent LLMs have been benchmarked in arithmetic tasks, including long-digits sum and multiplication, showing that LLMs such as ChatGPT and GPT-4 can perform reasonably well on these tasks even with no specific tuning. On the other hand, the accuracy of smaller models is markedly lower, and in general they are not able to work with long operands and generalize to OoD data.

Goat \citep{Liu2023} a LLaMA model specifically fine-tuned on arithmetic tasks performed even better than GPT-4 on large-number additions and subtractions, probably due to the consistent (digit level) tokenization of numbers in LLaMA models. However, it was able to perform multi-digits multiplication and division only forcing a Chain of Thought (CoT) \citep{Wei2023} decomposition of such tasks during instruction tuning.

\citet{Nogueira2021} tuned a T5-based pre-trained LM on additions and subtractions, and argued that tokenization and input representation are critical to achieve good accuracy. In particular, in their experiments character-based tokenization works better than sub-word tokenization, and making explicit the digit position in the input string (i.e., inserting after each digit a marker to denote its position in the sequence) generally leads to better accuracy. They also trained a vanilla non-pretrained LM on smaller numbers and found that classical sinusoidal-based positional embedding does not perform well, so they proposed a tailored position-wise masked embedding. Their paper contains other interesting findings such as the impact of the digit order (plain or reverse) and the size of the training set.

\citet{Muffo2023} tuned pre-trained GPT-2 models on 5-digit additions and 2-digit multiplications. They also found that making explicit the digit position in the input sequence helps to improve accuracy. While good accuracy is reported for addition, the tuned models struggle to learn multiplication even on two-digit operands.

\citet{Lee2023} trained small LMs to learn arithmetic tasks, mainly focusing on addition, but also experimenting with subtraction, multiplication, sine and square root. The authors carefully ablated different aspects of the training data to isolate the factors that contribute to the appearance of arithmetic capabilities. In particular, they studied the impact of the input order (plain or reverse) and the utility of providing intermediate information about the decomposition of the task in steps to promote CoT reasoning. Some results and findings included in \citet{Lee2023} will be further discussed throughout this paper.

All the above works provide useful contributions to understand the capabilities and limitations of large and small LMs to deal with arithmetic tasks, but none of them focus on the computational approach used to solve them, which is the main purpose of the present work (see Table \ref{table:InvestigationSteps}).

\subsection{Interpretability techniques}

A large number of techniques can be used to investigate the internal working mode of deep neural networks, including Transformers and LMs: see \citet{Rauker2023} for a recent survey. Weights, single neurons, subnetworks/circuits, and activations can be the target of \textit{intrinsic} approaches (implemented during training) or \textit{post-hoc} approaches (implemented after training).

Probing is a common technique used to investigate the representations learned by pre-trained LMs: it typically involves training a simple model (denoted as \textit{probe}) on top of the LM embeddings to predict a given property \citep{Belinkov2022}. Moreover, structural probing can be used to check whether internal representations encode discrete structures such as syntax trees \citep{Hewitt2019}, \citep{White2021}. However, a certain criticism emerged on probing analyses which is believed to disconnect the probing task from the original one and/or to reveal correlations instead of causations. Therefore, instead of focusing on the presence of information on internal encoding, some researchers proposed to check whether the removal of some knowledge from embeddings (e.g., amnesic probing \citep{Elazar2021}) negatively influences the model ability to perform a task \citep{Elazar2021}, \citep{Lasri2022}. Other interesting approaches to interpretability are mechanistic interpretability \citep{Elhage2021}  and causal abstraction \citep{Geiger2021}: the former is aimed at reverse engineering the algorithm that a model uses to solve a task and to map it to neural circuits; the latter constructs an interpretable causal model and aligns it with neural representations.

In this work we use a mix of intrinsic and post-hoc interpretability techniques: in particular through the experiments we manipulate the training set, change the input representation and the architecture components, perform correlation analyses of embeddings and apply amnesic probing.

\subsection{Interpretability of arithmetic reasoning with LMs} \label{subsec:MechanisticInterpretability}

\citet{Stolfo2023} introduced a causal mediation analysis to point out the LM components (e.g., attention heads, Multi-Layer Perceptrons - MLPs) involved in the information processing of simple arithmetic operations, focusing on the flow of numerical information throughout the model layers/columns. The main outcomes of this study are that the model: (i) processes the representation of numbers and operators with the first layers; (ii) information is then conveyed (by attention heads) to the last part of the sequence (i.e., output column), where (iii) it is numerically processed by late MLPs. 

\citet{Nanda2023} carefully studied the algorithmic approach put in place by a small Transformer to implement modular addition of small numbers. They discovered that the internal algorithmic implementation is based on discrete Fourier transforms and trigonometric identities to convert addition to rotation on a circle. While the outcomes are somewhat surprising, here the term algorithm must be taken with care: even if the experiments prove that internal processing well approximates given equations, the approach is a numerical approximation (based on weight encoded values) that does not generalize to different moduli (as a symbolic implementation of the equations could do).

Both these studies adopted a simplified setting where numbers are presented as single token, and the output is expected at the last position of the sequence. So the models are not operated in autoregressive manner and the multi-token encoding/decoding stages are simplified. In Section \ref{sec:Conclusions}  we discuss how the above findings are compatible with our findings.

\section{Experiment design} \label{sec:ExperimentDesign}

\subsection{The tasks} \label{subsec:TheTasks}

We focused on two simple computation tasks: binary addition and binary multiplication. Using binary encoding allows keeping the vocabulary very compact, since we need to encode only the symbols ‘0’, ‘1’ and a few other tokens. The selected tasks have other nice properties such as computing input similarities by Hamming distance and easily generating all combinations. Of course, a classical artificial neural network can be trained to learn to sum and multiply two integers or floating-point numbers, but adding/multiplying strings of tokens with an LM is trickier. 

More formally, given two integers $A$, $B$ (both in the range [0,127]) our input sequence (or prompt) is a 15-token string taking the form:

\begin{center}
$a_0 a_1 a_2 a_3 a_4 a_5 a_6$ $\langle op \rangle$ $b_0 b_1 b_2 b_3 b_4 b_5 b_6$
\end{center}

where $a_i,b_i\in\{\textrm{‘0’},\textrm{‘1’}\}$ are the symbols corresponding to bits in the \textit{i}-th position in the binary representation of $A$ and $B$ respectively, and $\langle op \rangle$ can be either ‘+’ or ‘×’.

The expected output string (or input completion) is:

\begin{center}
$R=r_0 r_1 ... r_{m-1}$
\end{center}

where $r_i$ is the \textit{i}-th bit in the binary representation of $A \langle op \rangle B$, and $m$ is the number of bits of the expected output string $R$ (8 and 14 for addition and multiplication, respectively).

It is worth noting that:
\begin{itemize}
\item we are using a fixed-length input/output representation (with zero padding for unused most significant bits) to make the digit positions more explicit.
\item in both the input and output the Least Significant Bits (LSBs) are provided before the Most Significant Bits (MSBs) (a.k.a., reverse or little-endian order) since this was supposed to simplify the model learning\footnote{in binary arithmetic the addition/multiplication algorithms start processing the LSBs in order to correctly propagate the intermediate carries.}. As discussed in \ref{sec:AppendixC} this assumption leads to a much faster training.
\end{itemize}

If we consider the sequence-to-sequence mapping underlying the proposed tasks we note that even in a simple binary addition a slight change in the input (i.e., a single bit) can produce a relevant change in the output because of the carries propagation. In the example below a single bit modification in the input produces an 8 bit modification in the output:

\begin{center}
$1000000+0111111 \rightarrow 11111110$

$1000000+1111111 \rightarrow 00000001$
\end{center}

Such input-output discontinuity is made more explicit for addition in \ref{sec:AppendixA}.

\subsection{The architecture} \label{subsec:TheArchitecture}

A non-pretrained encoder-decoder Transformer based on the original architecture introduced in \citet{Vaswani2017} was used as primary LM. Table \ref{table:LMDetails} reports the model setup and parametrization. The small vocabulary used allows us to keep the model small (just 701K learnable parameters) and trainable from scratch with a limited number of examples.

\begin{table}[h!]
\caption{Details of the LM model used in our experiments. The total number of learnable parameters is just 701K, which is several orders of magnitudes smaller than recent billion-parameters LLMs.}
\label{table:LMDetails}
\begin{center}
\begin{tabular}{ |c|c| } 
 \hline
 vocabulary size & 5 \\
 \hline
 vocabulary & {0: unused, 1: \textless start\textgreater, 2: ‘+’ or ‘×’, 3: ‘0’, 4: ’1’} \\ 
 \hline
 token embedding & learned \\
 \hline
 positional encoding & fixed (sinusoidal) \\
 \hline
 $d_{model}$ & 64 \\
 \hline
 $d_{ff}$ & $d_{model} \times 4$ \\
 \hline
 num\_heads \textit{h} & 8 \\
 \hline
 encoder layers & 6 \\
 \hline
 decoder layers & 6 \\
 \hline
 dropout & 0.1 \\
 \hline
 learnable parameters & 701K \\
 \hline
\end{tabular}
\end{center}
\end{table}

The LM was trained to learn separately the addition/multiplication tasks. For both problems, we exhaustively generated all the $2^{14}=16384$ input/output combinations, which were then randomly split into training ($3/4 \rightarrow 12288$) and validation ($1/4 \rightarrow 4096$) sets. In our experiments we do not need a separate dataset to tune hyperparameters so our validation set coincides with the test set.

An additional control experiment was run where the input sequences were the same of the addition experiment but the output completion was randomly generated (with the same length as the addition, i.e., 8 tokens). In this case, the lack of any dependencies between input and output makes it impossible to learn an algorithmic approach (or smooth mapping) to solve the problem and the only strategy to learn the training set is memorizing all the sequences.

When the trained LM is used in inference mode, we always pick the most probable token from the logit outputs (i.e., greedy decoding). Two metrics can be used to denote the LM accuracy: \textit{token accuracy} refers to the probability of generating the next token correctly, while \textit{sequence accuracy} refers to the probability of generating the whole output string correctly in autoregressive mode (i.e., generating one token at a time and appending it to the current prompt).

Most of the experiments have been repeated with a second LM (nanoGPT by \citet{Karpathy2022}) which is a good representative of the decoder-only family. Details are reported in \ref{sec:AppendixE}.

All the experiments included in this paper can be easily reproduced by running the code available at: (to be disclosed upon acceptance).

\section{Results} \label{sec:Results}

\subsection{Learning addition and multiplication} \label{subsec:LearningAdditionAndMultiplication}

\begin{figure}[h]
\begin{center}
\includegraphics[scale=1.05]{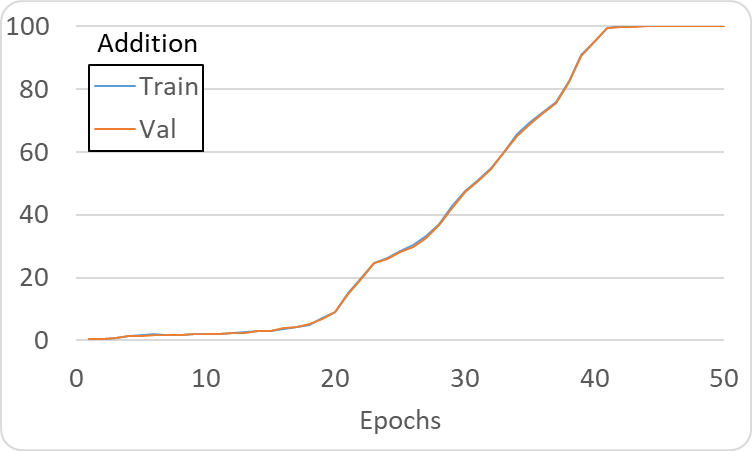}
\includegraphics[scale=1.05]{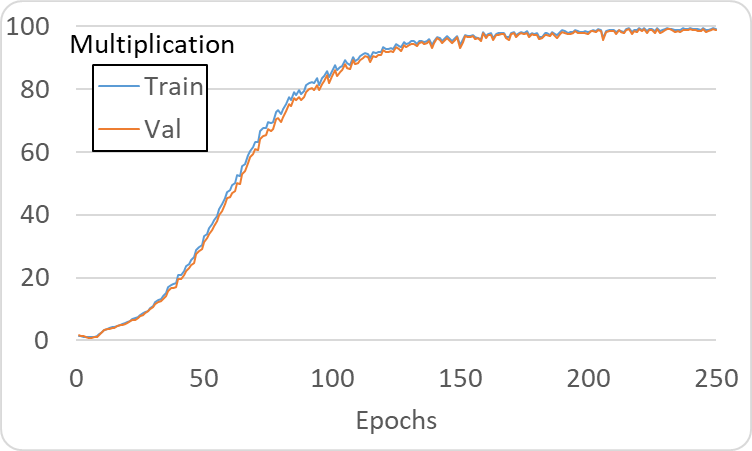}
\end{center}
\caption{Sequence accuracy. From the left: addition and multiplication. Results are averaged over five runs. Note that, training and validation curves are almost overlapped. At the end of training the Mean Absolute Error (MAE) on the validation set, between the real and generated operation results, is 0 and 1.3 for addition and multiplication, respectively.}
\label{figure:AddMulSeqAccuracies}
\end{figure}

Figure \ref{figure:AddMulSeqAccuracies} shows that our simple LM is able to learn addition in less than 50 epochs, and multiplication in about 250 epochs \footnote{We used the standard CrossEntropy loss, the Adam optimizer with the learning rate of 0.0001 and betas = 0.9 and 0.98, and a minibatch size of 128.}. As expected multiplication is more complex and requires more training: this is due to the high non-linearity of this operation (more on this later) and to the higher length of the output (14 vs 8 tokens). The accuracy on the validation set is very close to the training set, denoting almost perfect generalization on numbers never seen before. This is a somewhat surprising result, especially considering the limited size of the training data. No grokking\footnote{Grooking refers to the case where validation accuracy, much smaller than training accuracy at initial stages, suddenly increases after a certain number of epochs.} was observed \citep{Nanda2023}. Similar results were obtained with nanoGPT (see Figure \ref{figure:NanoGPTAddMulSeqAccuracies} in \ref{sec:AppendixE}.)

Unlike \citet{Nogueira2021} (see their Appendix B for a  similar setup), we were able to learn addition with the native sinusoidal positional encoding. Moreover, in \citet{Lee2023} additions can be effectively learnt by a simple LM, but to reach 100\% accuracy the training set had to be balanced in terms of the operand magnitude (i.e., number of digits) and carry propagation. The effectiveness of our training procedure is probably due to the lower complexity determined by a small vocabulary and fixed-length representation. As to multiplication, \citet{Muffo2023} were not able to effectively learn two (decimal) digits multiplication, while \citet{Lee2023} and \citet{Liu2023} had to provide extra intermediate steps in the prompt (denoted as \textit{detailed scratchpad}) or during instruction tuning, respectively. On the contrary our model effectively learnt multiplication of 7 binary digit operands: again the simplified setup may have been the key.

On the workstation used (with a single Titan RTX GPU) training can be completed in just 8 and 46 minutes for addition and multiplication, respectively. An estimation of the training complexity $C$ of an LLM  in term of floating point operations is $6\times N\times T$ \citep{Kaplan2020}, where $N$ is the number of model parameters (about 701K as reported in Table \ref{table:LMDetails}) and $T$ the number of training tokens. $T$ can be obtained as the product of the training set size (12288 in our experiments - see Section \ref{subsec:TheArchitecture}), the sequence length in tokens (23 and 29 for addition and multiplication, respectively - see Section \ref{subsec:TheTasks}) and the number of epochs (50 and 250 for addition and multiplication, respectively). Hence, for addition $T$ is 14M ($12288 \times 23 \times 50$) and therefore $C$ is about $59 \times 10^{12}$ operations while for multiplication $T$ is 89M ($12288 \times 29 \times 250$) and $C$ is about $374 \times 10^{12}$ operations.

\subsection{Control experiment: random output}

\begin{figure}[h]
\begin{center}
\includegraphics[scale=1.05]{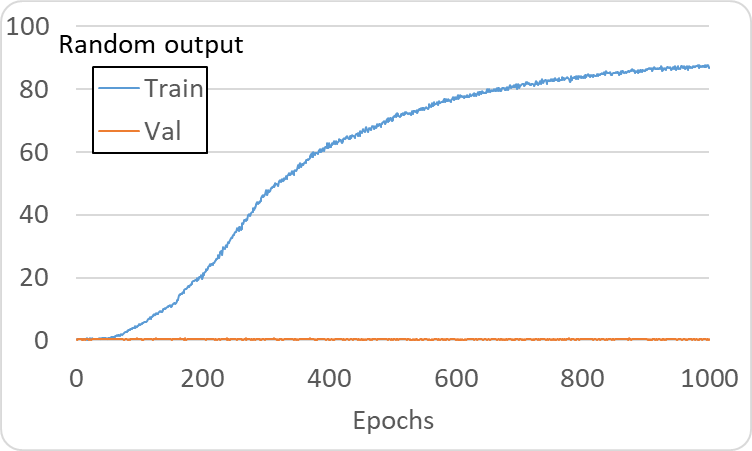}
\end{center}
\caption{Sequence accuracy using random output in the training set. Results are averaged over five runs.}
\label{figure:RandSeqAccuracies}
\end{figure}

If the output is randomly generated and therefore there is no relation with the input, the only possibility of learning the training set is by memorizing the whole data. Figure \ref{figure:RandSeqAccuracies} shows the training results: a much larger number of epochs (i.e., 1000) were necessary to reach a sequence accuracy of 87.8\%, and, as expected, the validation accuracy did not increase over the epochs. The difficulty of memorizing the training set (many more epochs) is due to the high discontinuity of the input-output mapping. In fact, because of the random output generation, very similar input sequences can be associated to completely different outputs.

Therefore, even if we only consider the accuracy on the training set, this result shows that an exhaustive memorization of the input is much more complex for the LM than solving the addition and multiplication tasks. This leads us to assume that, to efficiently solve the above computation tasks, the LM has found a computational approach (or algorithm) to simplify the output prediction. Now the question is: what is the approach? 

\subsection{The computational approach} \label{subsec:ComputationalApproach}

Let us consider two alternative approaches:

\textbf{Symbolic Manipulation (SM)}: a first idea is that the LM could learn the binary integer addition/multiplication algorithms used by an ALU inside a CPU (see \ref{sec:AppendixB} for a short reminder). Indeed, the addition algorithm is not complex and can be solved by using a 3-bit truth table (to sum each pair of corresponding bits with the carry-in) and iterative carry-out propagations. However, multiplication (by iterative additions) is much more complex and trickier to learn by using a symbolic manipulation approach. Furthermore, as shown in \citet{Lee2023}, a simple LM can also learn complex operations such as the sine function or the square root, whose mathematical (and algorithmic) decomposition is very complex since they require Taylor expansion and Newton method, respectively.

\textbf{Encoding-Regression-Decoding (ERD)}: if we consider the model architecture (Transformer) used for the LM and the underlying word embedding by vector representations, it is more likely that the LM solves the problem by decomposing it in the following three phases:

\begin{enumerate}
\item Encoding (token to value): maps the input sequence (i.e., $a_0 a_1 a_2 a_3 a_4 a_5 a_6$ $\langle op \rangle$ $b_0 b_1 b_2 b_3 b_4 b_5 b_6$) to a suitable vector representation. In principle, two vectors $\textbf{v}_A$ and $\textbf{v}_B$ representing the values (or magnitudes) of $A$ and $B$ are enough.
\item Regression: learns the computation as a supervised regression problem in the vector space: $\textbf{v}_R=regress(\textbf{v}_A,\textbf{v}_B)$. Actually this regression formulation is an oversimplification of the problem since in the next-token-prediction training the LM works incrementally. In \ref{sec:AppendixC} this discussion will be expanded. 
\item Decoding (value to token): maps the value vector $\textbf{v}_R$ back to token representation (i.e., $r_0  r_1...r_m$).
\end{enumerate}

It is worth noting that the above Encoding and Decoding phases do not need to be mapped onto the Transformer encoder and decoder (more on this later). The experiments reported in Sections \ref{subsec:InterpolationVsExtrapolation} and \ref{subsec:LookingAtInternalRepresentations} support the ERD assumption. The capability of capturing number magnitudes by pretrained embedders was also investigated by \citet{Wallace2019} who successfully trained a simple external regressor to compute the sum of two numbers starting from their embeddings. Other interesting studies on capturing numeracy with embedding were carried out by \citet{Naik2019} and \citet{Sundararaman2020}.

\subsection{Interpolation vs extrapolation} \label{subsec:InterpolationVsExtrapolation}

The random training/validation split performed for the experiments reported in Section \ref{subsec:LearningAdditionAndMultiplication} constitutes a somewhat simplified testbed to learn the two tasks. In fact, random split leads to a complete (even if sparse) coverage of the input space by both the training and validation sets, where each example in the validation set has high chance to be close to a training set example, and interpolation is enough to fill the gaps.

Hereafter, we exploit the well-known difficulty of a numerical regressor to work in the extrapolation regime to get insights about the computational approach of the LM. In particular, we considered two different criteria to isolate specific portion of the input space for the validation set, in order to better investigate extrapolation capabilities: 

\begin{itemize}
\item $VS_t=\{(A,B)|(A,B)\in NN_{4096}((A^*,B^*))\}$
                
        where $NN_{4096}((A^*,B^*))$ is the set of 4096 pairs $(A,B)$ which are the nearest neighbors to a centroid $(A^*,B^*)$ according to the Hamming distance between the corresponding token representations (i.e., number of different tokens at corresponding positions). As centroid $(A^*,B^*)$ in the token space we used: $1010101$ $\langle op \rangle$ $0101010$.
\item $VS_v=\{(A,B)|32 \leq A<96 \quad \textrm{and} \quad 32 \leq B<96\}$

        here the centroid is located in the middle of the value space (64, 64), so $VS_v$ is a squared region (of side 64) centered in the value space.
\end{itemize}

\begin{figure}[h]
\begin{center}
\includegraphics[scale=1.05]{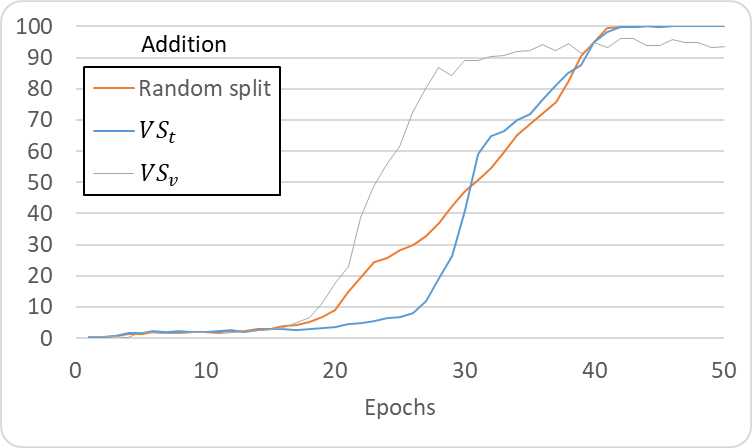}
\includegraphics[scale=1.05]{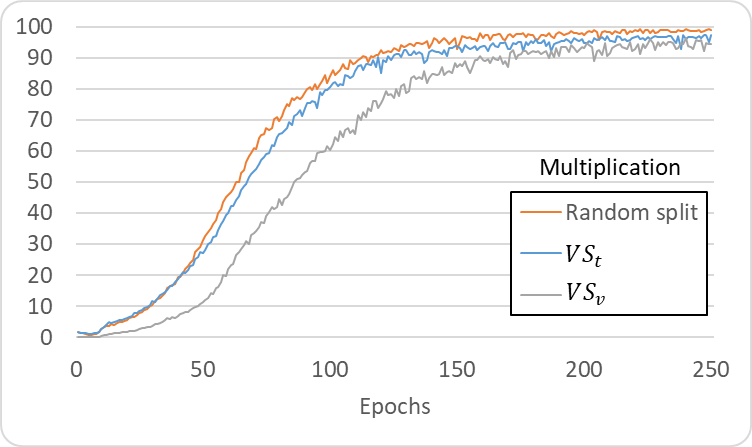}
\end{center}
\caption{Sequence accuracy on Random, $VS_t$, and $VS_v$ validation subsets for addition (left) and multiplication (right). Results are averaged over five runs. $VS_t$ reaches 100\% accuracy on additions (the same of Random split) and 97.5\% accuracy on multiplication (just 1.4\% less than random split); $VS_v$ reaches 93.7\% on addition and 94.3\% on multiplication (6.3\% and 4.6\% less than Random split, respectively).}
\label{figure:VStVSvSeqAccuracy}
\end{figure}

Both $VS_t$ and $VS_v$ isolate a contiguous data region of 4096 samples to be included in the validation set, but in the former the samples are close in the token representation space, while in latter are close in the value space. Being such contiguous portions of space excluded from the training set, we can expect a worse generalization. From the results (see Figure \ref{figure:VStVSvSeqAccuracy}) we note that $VS_t$ is very marginally affecting LM training and generalization while $VS_v$ has a major impact: in fact, in the second case, for both addition and multiplication the final sequence accuracy is from 4\% to 6\% points lower. This result strengthens the ERD hypothesis, since: (i) using $VS_v$ leads to the exclusion of a specific contiguous portion of value space during phase 2 and does not allow to properly train the regressor in this region; (ii) the encoding performed during phase 1 makes irrelevant the selection performed according to $VS_t$ because, after encoding, the corresponding data point remains scattered in the value space and the regressor can easily interpolate among them. Similar results were obtained with nanoGPT (see Figure \ref{figure:NanoGPTVStVSvSeqAccuracy} in \ref{sec:AppendixE}.)

\subsection{Looking at internal representations} \label{subsec:LookingAtInternalRepresentations}

Understanding the internal representation (embeddings in the vector space) in a trained Transformer is not an easy task. However, in the specific setting considered we can gain some hints by looking at the distances between the embedding of different data points (at different layers) and correlating them with the corresponding distances at input/output levels.

Given an LM trained on addition (or multiplication) we consider the dataset S including the 128 input pairs where the two operands have identical values\footnote{since the input prompt contains two operands, we select only the cases with identical values ($A=B$) in order to easily determine the “magnitude” of the input, and thereafter compute meaningful distances.}:

\begin{center}
$S=\{(A,A)|0 \leq A < 128$\}
\end{center}

At the input level (\textit{in}) we can compute two ordered sets of 8128 (128×127/2) distances each:

\begin{center}
$d_{in,t}=\{hdist(X,Y)|(X,X),(Y,Y) \in S,X<Y\}$
$d_{in,v}=\{|X-Y| \quad |(X,X),(Y,Y) \in S,X<Y\}$
\end{center}

where $hdist(X,Y)$ is the Hamming distance between the token representation of $X$ and $Y$, and the subscript letters $t$ and $v$ denote token and value levels, respectively.

At the output level (\textit{out}) we can compute the two corresponding sets of distances as:

\begin{center}
$d_{out,t}=\{hdist(P,Q)|(X,X),(Y,Y) \in S,X<Y\}$
$d_{out,v}=\{|P-Q| \quad |(X,X),(Y,Y) \in S,X<Y\}$
\end{center}

where ($P=X+X$ and $Q=Y+Y$) for addition, and ($P=X \times X$ and $Q=Y \times Y$) for multiplication.

Finally, for each intermediate level of the Transformer encoder (\textit{enc}) or decoder (\textit{dec}) we can compute the Euclidean distances among the corresponding embedding vectors. 

\begin{center}
$d_{enc_i}=\{\|enc_i(X,X)-enc_i(Y,Y)\| \quad |(X,X),(Y,Y) \in S,X<Y\}$
$d_{dec_i}=\{\|dec_i(X,X)-dec_i(Y,Y)\| \quad |(X,X),(Y,Y) \in S,X<Y\}$
\end{center}

where $enc_i$ and $dec_i$ are the output vectors obtained by concatenating all the token embeddings (each of dimensionality 64) after the \textit{i}-th encoder and decoder layer, respectively. For example $enc_i$ has dimensionality $960 = 64 \times 15$ where 15 is the number of tokens in the encoder.

\begin{figure}[!h]
\begin{center}
\includegraphics[scale=1.2]{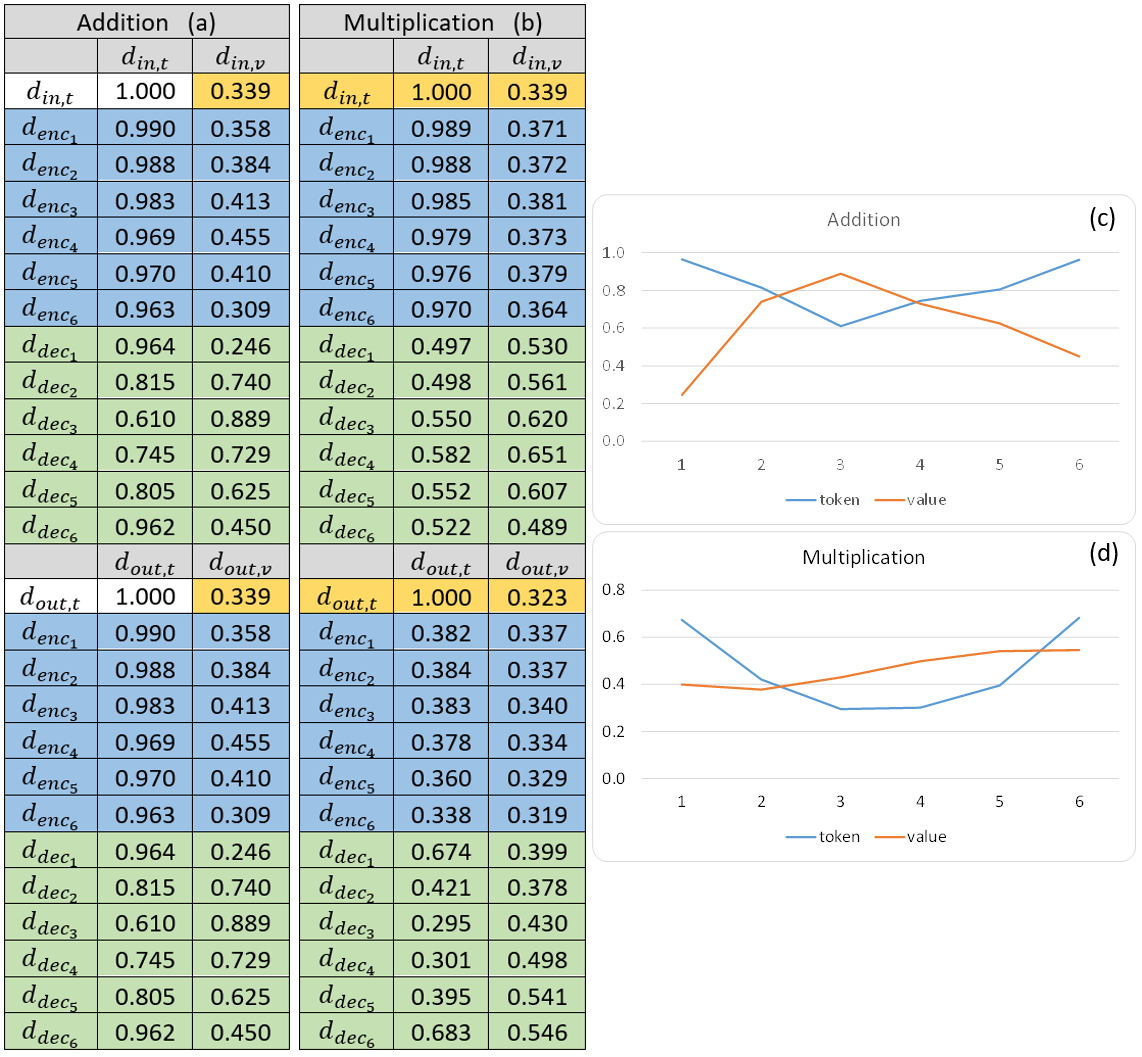}
\end{center}
\caption{Pearson correlation between ordered sets of distances for addition (a) and multiplication (b). Each cell denotes the correlation between the two ordered set of distances specified in the corresponding row and column. Note that since for addition in this experiment the output value is always twice the input, the correlation values (blue and green cells) are the same for $d_{in,\_}$ and $d_{out,\_}$ block of values. Graphs (c) and (d) show the correlations of output distances $d_{out,t}$ (at token level - blue curves) and $d_{out,v}$ (at value level - orange curves) with the embedding distances $d_{dec_i}$ across the 6 decoder layers for addition and multiplication, respectively.}
\label{figure:PearsonCorrelation}
\end{figure}

Even if the distances in the different sets have different ranges, we can use correlation to find out similarities. If two sets of distances are correlated we can expect that the corresponding representations/embeddings are correlated as well. Since both Pearson and Spearman correlations \citep{Schober2018} provided similar outputs, for simplicity in Figure \ref{figure:PearsonCorrelation} we report only Pearson correlations. 

The yellow cells in the tables of Figure \ref{figure:PearsonCorrelation} confirm the low correlation between the token and value representation at both input and output level. The blue cells show that correlation remains quite similar across the encoder layers as if the encoder was not performing any significant computation (this is confirmed in Section \ref{sec:AblationStudy} where we achieve similar results by totally removing all intermediate attention and MLP layers in the encoder). More interesting is the trend of correlations across the decoder layers (green cells). In particular, for the addition the token representation has high correlation with the first and last layers and low with central layers, while the value representation has an opposite trend (see also Figure \ref{figure:PearsonCorrelation}.c). These results support the ERD hypothesis and in particular that the initial and final layers in the decoder transform from token to value representation (and vice versa) while the central layers perform regression in the value space. In particular, at layer 3, the correlation at token level is minimum while the correlation at value level is maximum.

For multiplication the low-high-low trend at value level is less evident (Figure \ref{figure:PearsonCorrelation}.d orange curve), probably because the quadratic dependence of the output from the input (at value level) does not allow to learn a simple regressor smoothly working in the whole vector space, and the mapping is performed by piecewise linear approximation in different space regions, which introduces discontinuities that make global distances in the vector space unsuitable to quantify the representation similarity.

As discussed in Section 2.2, correlation analyses might be insufficient to prove that the presence of a certain information in the embeddings is really necessary to compute the output (direct causation). So to further strengthen our hypothesis we applied an amnesic probing technique \citep{Elazar2021} and proved that, upon removal of value information from the embeddings, the LM is no longer capable of performing the right computation. Details are reported in Appendix D.  

\section{Ablation study} \label{sec:AblationStudy}

This section presents the results of an ablation study where the LM architecture was simplified, to understand which components are necessary to learn the addition/multiplication computation. Squeezing the encoder (i.e., removing all intermediate attention and MLP layers) does not have a relevant impact; this is consistent with other works claiming that a decoder only architecture \citep{Liu2018} can achieve similar results with respect to an encoder-decoder Transformer, and further confirmed by the nanoGPT results presented in \ref{sec:AppendixE}. A simplification of the architecture in terms of (i) reduction of dimensionality; (ii) reduction of number of heads; (iii) removal of fully connected layers is well tolerated, while positional embedding and attention layers are mandatory for the LM in order to properly perform token to value transformation (and vice versa).
Table \ref{table:AblationStudy} summarizes the results.
\begin{table}[h]
\caption{Epochs necessary to reach 95\% accuracy on the validation set. A dash is used when 95\% accuracy is not achieved in 1K epochs: in such case the accuracy reached is reported within brackets.}
\label{table:AblationStudy}
\begin{center}
\begin{tabular}{ |c|c|c| } 
 \hline
 \textbf{Configuration} & \textbf{Addition} & \textbf{Multiplication} \\
 \hline
 Full (see Table \ref{table:LMDetails}) & 39 & 137 \\ 
 \hline
 Squeezing the encoder (see main text) & 60 & 426 \\
\hline
 num\_heads \textit{h}=1 & 25 & 225 \\
 \hline
 Reduced dimensionality ($d_{model}=32$) & 66 & 309 \\
 \hline
 No positional embedding & --- (2.4\%) & --- (1.8\%) \\
 \hline
 No attention layers & --- (0.9\%) & --- (1.7\%) \\
 \hline
 No fully connected layers & 56 & 398 \\
 \hline
\end{tabular}
\end{center}
\end{table}

\section{Discussion and conclusions} \label{sec:Conclusions}

In this paper we introduced a simplified setup to allow a light LM to learn binary addition and multiplication. Both the LM architectures considered easily learn the two tasks and generalizes well on unseen data, proving that memorization of the training data is neither necessary nor efficient. The experiments on the interpolation/extrapolation capabilities and correlation of input-output representations with internal embedding suggest that the model solves the computational task as a supervised regression problem in the value space after an initial encoding from token to values, and a final decoding from output value to tokens. Under this hypothesis: (i) any task that can be solved by a neural network regressor can be solved by an LM as well, with the extra burden of end-to-end learning decoding/encoding steps; (ii) when looking at interpolation/extrapolation capabilities of an LM applied to a mathematical task, we should not concentrate on the input token representation but on the internal representation after encoding, keeping in mind the difficulties of a numerical regressor to work on region spaces not covered by the training set; (iii) on a more speculative side, we could guess that modern LLMs learn the number encoding/decoding once and reuse it across different numerical tasks whereas a specific regressor is learned for each task.

Our ERD hypothesis could be questioned considering some recent findings from \citet{Lee2023} where providing in the prompt intermediate information (scratchpad) about the decomposition of arithmetic tasks improves the training efficiency and requires fewer examples. This could suggest that a symbolic manipulation approach is adopted to learn imitating step by step the proposed decomposition. However, in most of the cases their model was able to learn the same task (even if slowly) without scratchpad and/or with wrong scratchpads. As argued by the authors the higher efficiency is actually in terms of examples and not in terms of tokens since each scratchpad requires a large number of extra tokens, and we guess these could be used as extra features by the underlying regressor. Furthermore, scratchpad contribution is negligible for more complex operations such as sine and square root, but, unexpectedly, learning such complex operations was simpler than multiplication. This is not strange under the ERD hypothesis where a unary smooth operator like the sine can be learned by a supervised regressor independently of the mathematical method used for its computation.

The algorithmic interpretation that \citet{Nanda2023} provided for modular addition, could also suggest that their LM discovered and efficient symbolic manipulation approach; however, as discussed in Section \ref{subsec:MechanisticInterpretability}, it is more likely that a regressor was learned to numerically approximate an efficient sparse Fourier decomposition, under regularization constraints favoring sparsity. Finally, the information flow described in \citet{Stolfo2023}, points out that MLPs in the last layers are responsible for the numerical computation of the solution, which is compatible with the hypothesis of a multi-layer regressor.

Of course we are not claiming that all the capabilities of modern LLMs can be explained by regression, but regression is likely to be one of the internal tools that LLMs uses to predict the next token when numbers come into play. 

As to future research we plan to: i) further investigate the generalization capabilities of LMs in arithmetic tasks with respect to the composition of the training and test sets \citep{Feng2023,Keskar2017}, ii) design simplified experiments/setups for tasks that cannot be easily mapped to regression problems such as chain of reasoning and logic deductions.

\appendix

\section{Addition input-output discontinuities} \label{sec:AppendixA}

Given an input/output pair we consider the ($2^{14}$) variants obtained by perturbing (i.e., 0-1 swap) the input bits and counting the resulting changes in the output. These values, averaged over all possible input/output pairs (again $2^{14}$) and normalized by row are inserted in the cells of \ref{table:AdditionDiscontinuities}. So, for example the value in cell (row=2, column=3) means that in the 27.9\% of the cases a perturbation of 2 (over 14) bits in the input leads to a change of 3 (over 8) bits in the output.

\begin{table}[h!]
\caption{Addition input-output discontinuities.}
\label{table:AdditionDiscontinuities}
\includegraphics[width=\textwidth]{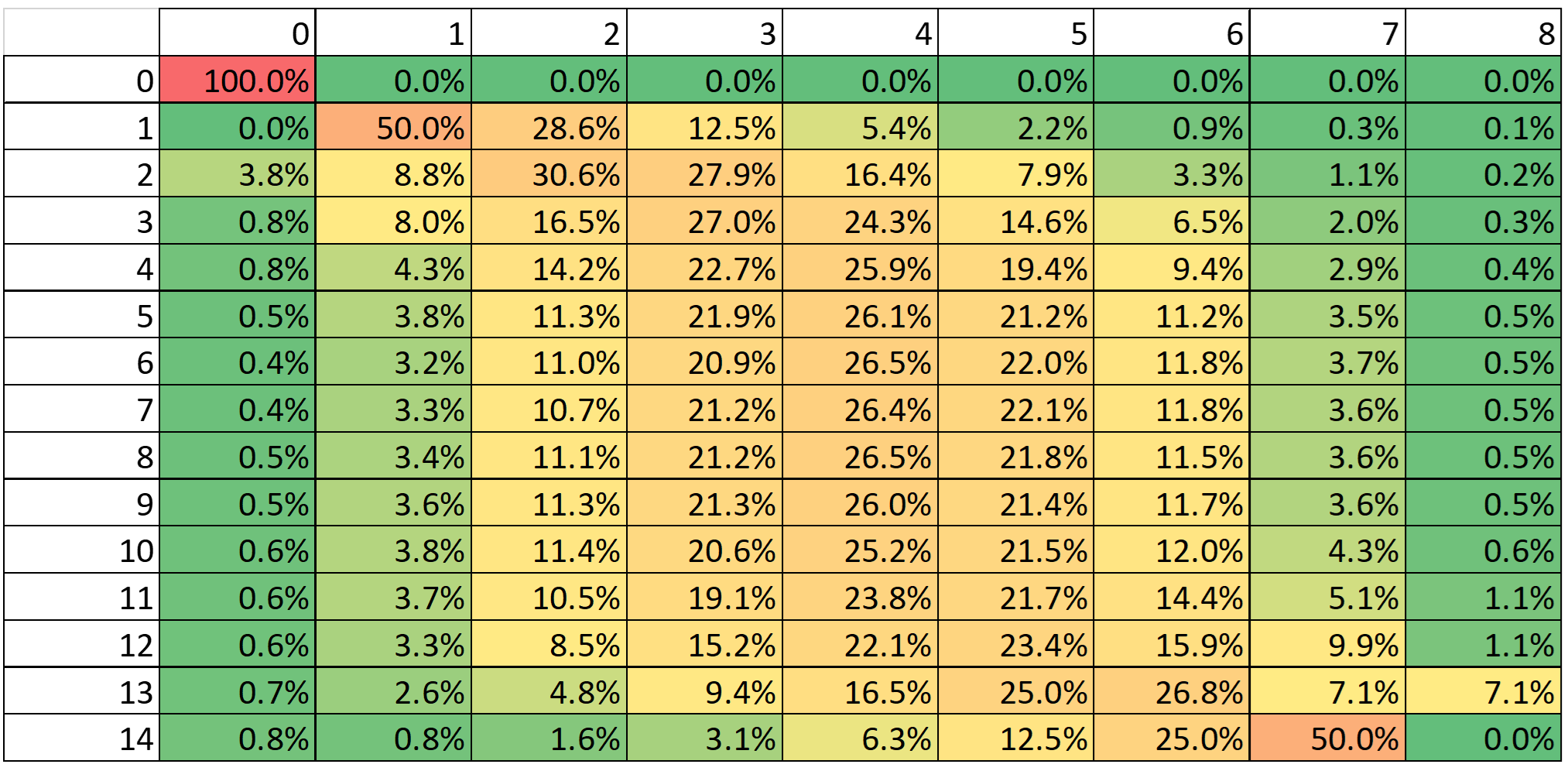}
\end{table}

Input-output discontinuities, which are further amplified in case of multiplications, make it very unlikely to solve these tasks by smooth interpolation of the input representation.

\section{Binary addition and multiplication} \label{sec:AppendixB}

Binary addition can be executed by summing pairs of corresponding bits $a_i$ and $b_i$, starting from the LSBs ($a_0$ and $b_0$) and propagating carries. Let $c_{i-1}$ be the pending carry used to sum current bits \footnote{when summing the LSBs ($i=0$), there is no pending carry, so $c_{-1}=0$}, then a two-output 3-bit truth table (Table \ref{table:TwoOutputTruthTableForAddition}) can be used to generate the output bit $o_i$ and carry $c_i$ used when summing the next pair of bits:

\begin{table}[h!]
\caption{Two-output 3-bit truth table for binary addition.}
\label{table:TwoOutputTruthTableForAddition}
\begin{center}
\begin{tabular}{ |c|c|c|c|c| } 
 \hline
 \multicolumn{3}{|c|}{Inputs} & \multicolumn{2}{c|}{Outputs}  \\
 \hline
 $a_i$ & $b_i$ & $c_{i-1}$  & $o_i$ & $c_i$ \\ 
 \hline
 0 & 0 & 0 & 0 & 0 \\ 
 \hline
 0 & 0 & 1 & 1 & 0 \\ 
 \hline
 0 & 1 & 0 & 1 & 0 \\ 
 \hline
 0 & 1 & 1 & 0 & 1 \\ 
 \hline
 1 & 0 & 0 & 1 & 0 \\ 
 \hline
 1 & 0 & 1 & 0 & 1 \\ 
 \hline
 1 & 1 & 0 & 0 & 1 \\ 
 \hline
 1 & 1 & 1 & 1 & 1 \\ 
 \hline
 \end{tabular}
\end{center}
\end{table}

A simple approach to execute binary multiplication is through iterative binary sums. Each bit $b_i$ of the second operand is multiplied by the whole first operand, but this inner multiplication is straightforward since it results either in a sequence of 0 (if $b_i=0$) or a copy of the first operand (if $b_i=1$). This intermediate result is then shifted left and summed to the current output. An example is reported in Figure \ref{figure:ExampleMultiplication} below.

\begin{figure}[!h]
\begin{center}
\includegraphics[scale=1.2]{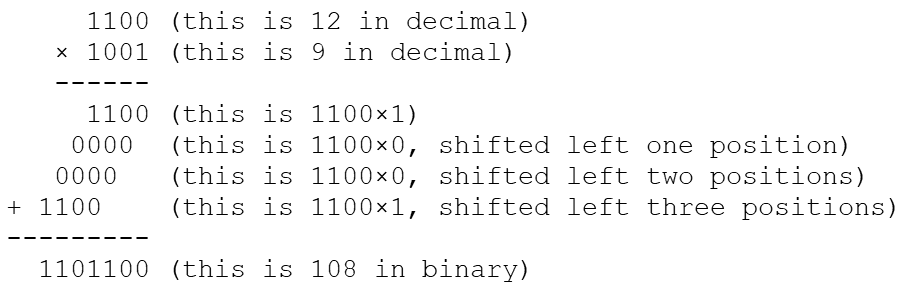}
\end{center}
\caption{Example of 4-digit binary multiplication. The sum can be performed incrementally with a two-operand adder.}
\label{figure:ExampleMultiplication}
\end{figure}

\section{Learning a regressor under predict-next–token training} \label{sec:AppendixC}

In Section \ref{subsec:ComputationalApproach} we argued that an arithmetic computation task can be decomposed into three steps whose central one is learning a regressor in the value space: $\textbf{v}_R=regress(\textbf{v}_A,\textbf{v}_B)$. If we consider the autoregressive working mode of a Transformer and its predict-next-token training, the regressor must be able to work incrementally given the output produced so far. In particular, we can formulate the problem as: $\textbf{v}_{r_i}=regress(\textbf{v}_A,\textbf{v}_B,i,\textbf{c}_{R_{i-1}})$ where:

\begin{itemize}
\item $\textbf{v}_A=[\textbf{v}_{a_0} \textbf{v}_{a_1}...\textbf{v}_{a_6}]$ and $\textbf{v}_B=[\textbf{v}_{b_0} \textbf{v}_{b_1}...\textbf{v}_{b_6}]$ are the value vectors of the two input operands, obtained as the concatenation of the value vectors of single tokens. Both are always fully available to the decoder. Note that, $\textbf{v}_{a_i}$ and $\textbf{v}_{b_i}$ are not the bits of the inputs, but correspond to their value vectors including also positional information.

\item $i$ is the position of the token to be predicted (we can assume it is available through positional encoding).

\item $\textbf{c}_{R_{i-1}}=[\textbf{c}_{r_0}\textbf{c}_{r_1}...\textbf{c}_{r_{i-1}}]$ is a value vector encoding the current context determined by the result produced so far (entering in the decoder from the bottom).

\item $\textbf{v}_{r_i}$ is the value vector of the $i$-th token. 
\end{itemize}

In principle, the regressor could predict each $\textbf{v}_{r_i}$ based on $\textbf{v}_A$ and $\textbf{v}_B$ alone, but we argue that the exploitation of the result produced so far $\textbf{c}_{R_{i-1}}$ can lead to higher training efficiency. To this purpose is interesting to evaluate the impact of the output ordering (plain or reverse). In both the addition and multiplication the $i$-th token of the result only depends on the tokens of the inputs at positions $\leq i$ (see \ref{sec:AppendixB}). Therefore, if reverse order is adopted, as we assumed until now, $\textbf{v}_{A_i}=[\textbf{v}_{a_0} \textbf{v}_{a_1}...\textbf{v}_{a_i}]$, $\textbf{v}_{B_i}=[\textbf{v}_{b_0} \textbf{v}_{b_1}...\textbf{v}_{b_i}]$ and $\textbf{c}_{R_{i-1}}$ are sufficient to predict $\textbf{v}_{r_i}$. Viceversa, if the output computation starts with the MSBs the regressor cannot leverage the above iterative decomposition and needs to learn the task as a global operation using whole vectors $\textbf{v}_A$ and $\textbf{v}_B$ with almost no support from the result produced so far.

\begin{figure}[h]
\begin{center}
\includegraphics[scale=1.05]{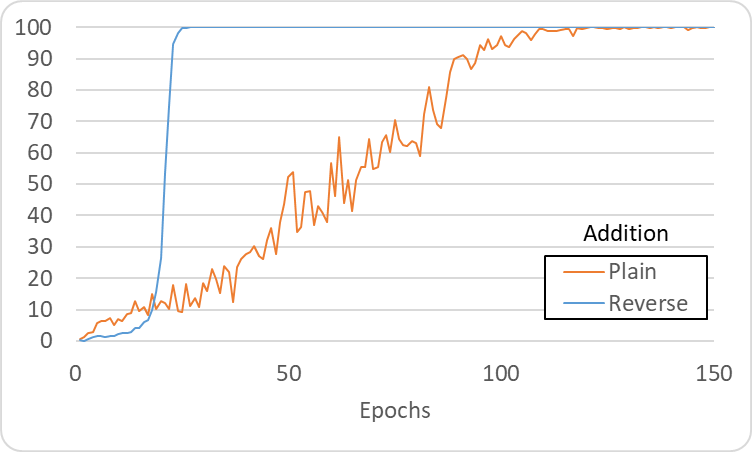}
\includegraphics[scale=1.05]{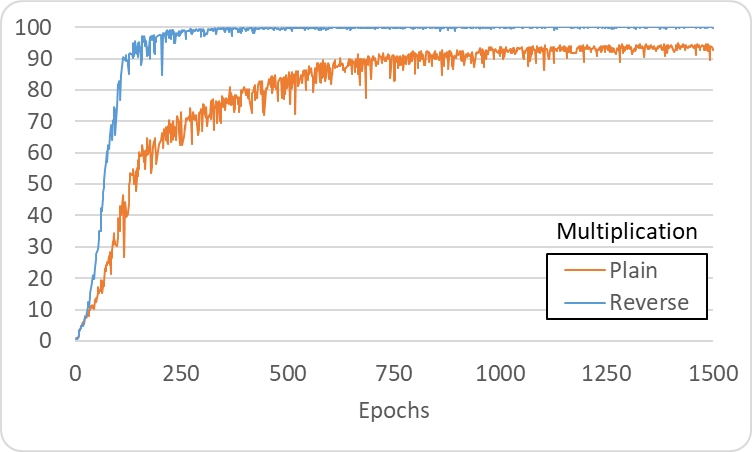}
\end{center}
\caption{Sequence accuracy on validation set for reverse (default in this work) and plain order of the input and output representations. From left to right: addition and multiplication.}
\label{figure:PlainRevBitsSeqAccuracies}
\end{figure}

In Figure \ref{figure:PlainRevBitsSeqAccuracies} we note that with plain order both addition and multiplication require a much longer number of epochs to converge and the learning curve is less stable. Further experiments proved that, as expected, the order of the inputs (also reverse by default in this study) is irrelevant, since the LM can always access the whole input representations $\textbf{v}_A$ and $\textbf{v}_B$. The advantages of using the reverse order are pointed out in other recent studies \citep{Nogueira2021,Lee2023}. In particular, \citet{Lee2023} reported a significant improvement with respect to plain order.

\section{Amnesic probing results} \label{sec:AppendixD}

The outcome of correlation analyses performed in Section \ref{subsec:ComputationalApproach} suggests that the embeddings in the central layers of the decoder contain information related to the value representation of the output (see Figure \ref{figure:PearsonCorrelation}). However correlation does not mean causation, and here we investigate deeper. Amnesic probing was proposed in \citet{Elazar2021} building on the approach \citet{Ravfogel2020} to check to what extent a model output is affected by the removal of specific features or attributes in an intermediate level embeddings. Here we focus on addition and we try to remove some features from the decoder layer 3 embeddings ($dec_3(X,Y)$). To this purpose a linear probe (a linear regressor in our case) was trained to predict the output value ($X+Y$) starting from the $dec_3(X,Y)$ embeddings and its nullspace is used to project the embeddings in a new space lacking output value information. According to \citet{Ravfogel2020}, due to the simplicity of the linear regressor used, the procedure is repeated twice to remove more information. Our results show that:
\begin{enumerate}
\item A simple linear regressor trained on $dec_3(X,Y)$ embeddings can reach high accuracy in predicting $X+Y$ (rmse = 0.28).
\item If the projected embeddings are overwritten in the LM decoder at level 3, and a partial forward pass is performed thereafter, the addition sequence accuracy severely drops from 100\% to 0.13\%.
\item As indicated in \citet{Elazar2021} since any information removal could hamper the model accuracy, a control test was performed by removing the same amount of information (but on random directions instead of the nullspace directions) and in this case the LM final sequence accuracy remained 100\%. 
\end{enumerate}
This experiment provides further support to the hypothesis that the value information is not only present in the inspected embeddings but is also crucial for the output computation.

On the computational side, we argue that amnesic probing complexity is low because it relies on simple steps as linear regression and null space computation, with the former being the most demanding step. Linear regression complexity is $O(nd^2+d^3)$ where $n$ is the number of training examples and $d$ the dimensionality of the embeddings.

\section{NanoGPT - a decoder-only LM} \label{sec:AppendixE}

To demonstrate that our findings generalize beyond the encoder-decoder architecture of the original Transformer used in this work, the main experiments have been repeated using a second LM, that is the nanoGPT \citep{Karpathy2022} decoder-only model. Table \ref{table:NanoGPTDetails} reports the details of the nanoGPT model adopted.

\begin{table}[h!]
\caption{Details of the nanoGPT model.}
\label{table:NanoGPTDetails}
\begin{center}
\begin{tabular}{ |c|c| } 
 \hline
 token embedding & learned \\
 \hline
 positional encoding & learned \\
 \hline
 $d_{model}$ & 64 \\
 \hline
 $d_{ff}$ & $d_{model} \times 4$ \\
 \hline
 num\_heads \textit{h} & 8 \\
 \hline
 decoder layers & 6 \\
 \hline
 dropout & 0.1 \\
 \hline
 learnable parameters & 298K \\
 \hline
\end{tabular}
\end{center}
\end{table}

\begin{figure}[h]
\begin{center}
\includegraphics[scale=1.05]{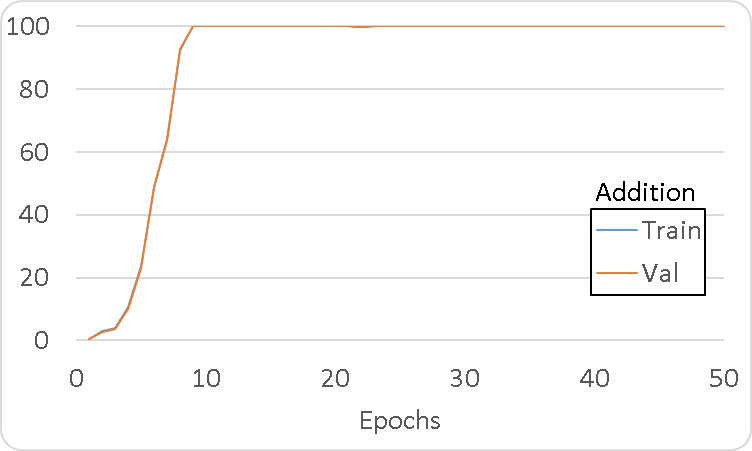}
\includegraphics[scale=1.05]{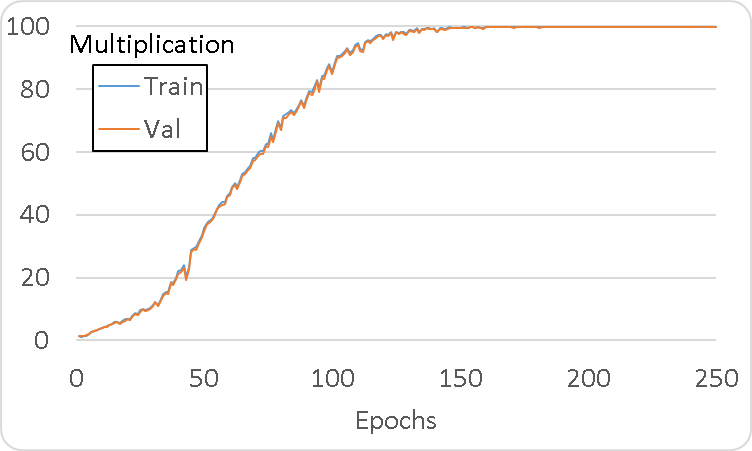}
\end{center}
\caption{Sequence accuracy of the nanoGPT model (refer to Section \ref{subsec:LearningAdditionAndMultiplication} for more details). From the left: addition and multiplication. Results are averaged over five runs. Note that, training and validation curves are almost overlapped.}
\label{figure:NanoGPTAddMulSeqAccuracies}
\end{figure}

Figure \ref{figure:NanoGPTAddMulSeqAccuracies} shows that the nanoGPT model was able to learn addition and multiplication still more efficiently than the original Transformer (compare Figure \ref{figure:AddMulSeqAccuracies} with Figure \ref{figure:NanoGPTAddMulSeqAccuracies}). For the training, we used a minibatch size of 128, a standard CrossEntropy loss, the AdamW optimizer with a learning rate of 0.001 and betas = 0.9 and 0.98, and a gradient clipping to 1.0.

\begin{figure}[h]
\begin{center}
\includegraphics[scale=1.05]{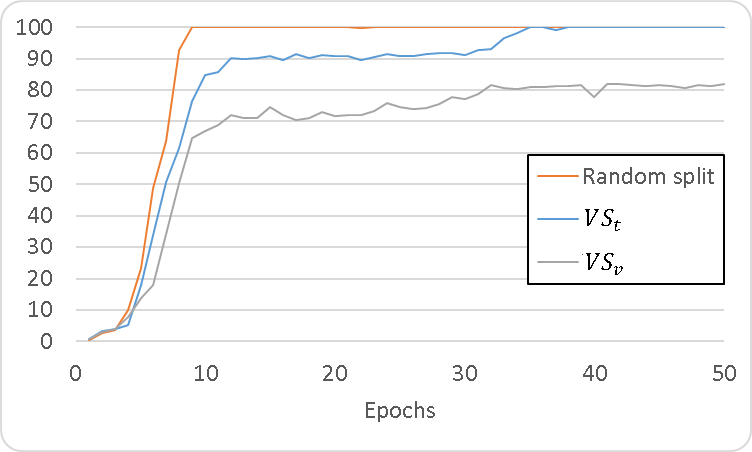}
\includegraphics[scale=1.05]{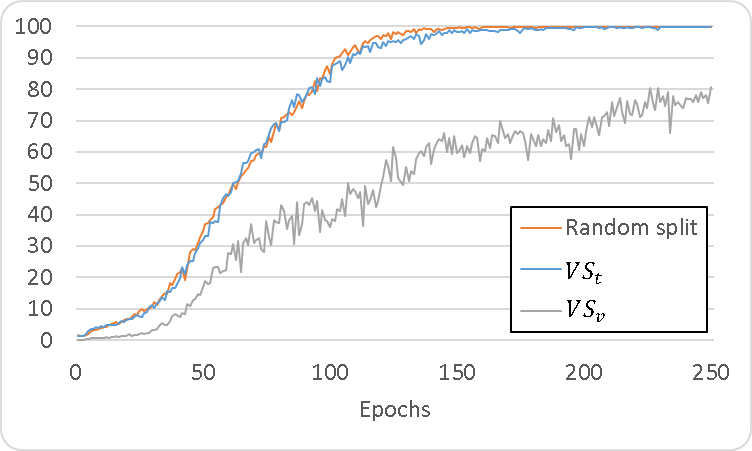}
\end{center}
\caption{Sequence accuracy of the nanoGPT model on Random, $VS_t$, and $VS_v$ validation subsets for addition (left) and multiplication (right). Results are averaged over five runs.}
\label{figure:NanoGPTVStVSvSeqAccuracy}
\end{figure}

Figure \ref{figure:NanoGPTVStVSvSeqAccuracy} shows the sequence accuracy of the nanoGPT model on Random, $VS_t$, and $VS_v$ validation subsets for addition and multiplication (see Section \ref{subsec:InterpolationVsExtrapolation} for more details). Using the $VS_t$ subset, it reaches 100\% and 99.9\% accuracy on addition and multiplication, respectively (the same of Random split) while, using the $VS_v$ subset, it reaches 82.0\% on addition and 80.6\% on multiplication (18.0\% and 19.4\% less than Random split, respectively). Results are inline with those obtained in Section \ref{subsec:InterpolationVsExtrapolation} but here the difference between $VS_t$, and $VS_v$ is still more significant.

\end{document}